\documentclass[runningheads, a4paper]{llncs}
\usepackage{amssymb, amsmath, bm, latexsym, comment}
\usepackage{xcolor}
\usepackage{graphicx}
\usepackage[subrefformat=simple,labelformat=simple]{subcaption}
\usepackage{indentfirst}
\usepackage{setspace}
\usepackage{verbatim}
\usepackage{array}
\usepackage{cite}
\usepackage{booktabs}
\usepackage{hyperref}
\usepackage[capitalize]{cleveref}
\usepackage{arydshln}
\usepackage{multirow}
\usepackage{dsfont}
\usepackage{mathtools}

\usepackage{booktabs} % for professional-looking tables
\usepackage{multirow} % for multi-row cells

\usepackage{booktabs} % for professional-looking tables
\usepackage{colortbl} % for coloring table cells
\definecolor{lightgray}{gray}{0.9} % define a light gray color

\usepackage{amsfonts}
\usepackage{siunitx}
\usepackage[misc,geometry]{ifsym}

\newcolumntype{C}{>{\centering\arraybackslash}X}
\newcolumntype{L}{>{\raggedright\arraybackslash}X}
\newcolumntype{R}{>{\raggedleft\arraybackslash}X}

\usepackage{physics}
\usepackage{environ}
\usepackage[linesnumbered,ruled,vlined]{algorithm2e}
\SetAlFnt{\small}
\usepackage{algpseudocode}
\usepackage{nicefrac}
\crefname{algocf}{Algorithm}{Algorithms}

\graphicspath{{./figures/}}

\newcolumntype{L}[1]{>{\raggedright\let\newline\\\arraybackslash\hspace{0pt}}m{#1}}
\newcolumntype{C}[1]{>{\centering\let\newline\\\arraybackslash\hspace{0pt}}m{#1}}
\newcolumntype{R}[1]{>{\raggedleft\let\newline\\\arraybackslash\hspace{0pt}}m{#1}}

\DeclarePairedDelimiterX{\infdivx}[2]{\big[}{\big]}{%
  #1\;\delimsize\|\;#2%
}

\DeclareMathOperator*{\argmax}{arg\,max}
\DeclareMathOperator*{\argmin}{arg\,min}

\newlength{\myl}
\let\origequation=\equation
\let\origendequation=\endequation
\RenewEnviron{equation}{
  \settowidth{\myl}{$\BODY$}                       % calculate width and save as \myl
  \origequation
  \ifdimcomp{\the\linewidth}{>}{\the\myl}
  {\ensuremath{\BODY}}                             % True
  {\resizebox{\linewidth}{!}{\ensuremath{\BODY}}}  % False
  \origendequation
}

\makeatletter
\newcommand{\printfnsymbol}[1]{%
  \textsuperscript{\@fnsymbol{#1}}%
}
\makeatother

\begin{document}

% \mainmatter  % start of an individual contribution
\title{Toward Universal Medical Image Registration via Sharpness-Aware Meta-Continual Learning}

\titlerunning{Toward Universal Medical Image Registration}

\author{
Bomin Wang\thanks{Equal contribution. \Letter\, Corresponding author.}
\and
Xinzhe Luo\printfnsymbol{1}
\and
Xiahai Zhuang\textsuperscript{\Letter} 
} %1{Wang, Bomin}, %2{Luo, Xinzhe}, %3{Zhuang, Xiahai}

\authorrunning{B. Wang, et al.}   % abbreviated author list (for running head)

% list of authors for the TOC (use if author list has to be modified)
\tocauthor{}

\institute{School of Data Science, Fudan University, Shanghai, China \\
\email{bmwang21@m.fudan.edu.cn, \{xzluo19, zxh\}@fudan.edu.cn}}

%\toctitle{Lecture Notes in Computer Science}
%\tocauthor{Authors' Instructions}
\maketitle

\begin{abstract}
Current deep learning approaches in medical image registration usually face the challenges of distribution shift and data collection, hindering real-world deployment. In contrast, universal medical image registration aims to perform registration on a wide range of clinically relevant tasks simultaneously, thus having tremendous potential for clinical applications. In this paper, we present the first attempt to achieve the goal of universal 3D medical image registration in sequential learning scenarios by proposing a continual learning method. Specifically, we utilize meta-learning with experience replay to mitigating the problem of catastrophic forgetting. To promote the generalizability of meta-continual learning, we further propose sharpness-aware meta-continual learning (SAMCL). We validate the effectiveness of our method on four datasets in a continual learning setup, including brain MR, abdomen CT, lung CT, and abdomen MR-CT image pairs. 
Results have shown the potential of SAMCL in realizing universal image registration, which performs better than or on par with vanilla sequential or centralized multi-task training strategies.
The source code will be available from \url{https://github.com/xzluo97/Continual-Reg}. 
\end{abstract} 

\section{Introduction}\label{sec:introduction}
Medical image registration aims to estimate the optimal
spatial transformation to align the structures of interest between a pair of fixed and moving images, which is a fundamental task in medical image analysis and has been widely used in disease monitoring, surgical navigation, and image diagnostics \cite{backgupta2021study, backmaes1997multimodality, backmaintz1998survey}. Over the past decade, advancements in artificial intelligence (AI) and deep learning (DL) technologies have significantly transformed the landscape of medical image registration. DL-based methods have shown promising results in enhancing the accuracy and efficiency of image registration \cite{backhaskins2020deep}. However, two challenges exist that hinder the real-world applications of DL-based methods.

\textbf{The challenge of distribution shift.} Due to the data-dependent nature of DL algorithms, their performance significantly depends on how the data are generated in speciﬁc contexts, at speciﬁc times. Therefore, most current learning-based medical image registration methods are task-specific (e.g., anatomy- and appearance-specific), and thus can be performed only on data belonging to the distribution of a single registration task\cite{backkang2022dual, backche2023amnet, backhering2021cnn}. When dealing with out-of-distribution data, significant performance decay will occur. 

\textbf{The challenge of centralized data collection.} In highly dynamic environments like healthcare, data are typically sequentially collected over time. In addition, the issue of data privacy makes it challenging to centralize the medical data. Therefore, it is difficult to collect all data simultaneously to train a universal registration model for different tasks. Meanwhile, the problem of distribution shifts occurs when sequentially learning on different datasets using current DL methods, which can lead to catastrophic forgetting of previously learned tasks, that is the performance degrades significantly after learning new data. 

Currently, DL-based algorithms cannot be modiﬁed after approval for clinical use \cite{backvokinger2021continual, backlee2020clinical}. These locked and static AI and DL systems are constrained in their capacity to learn from post-approval, real-world clinical data, thereby precluding the potential for achieving universal registration models. Continual learning (CL) methods aim to endow the ability to learn continuously to DL models while reducing forgetting \cite{journal/pnas/kirkpatrick2016, conference/iclr/saha2021, conference/neurips/rolnick2018}. 

\begin{figure}
    \centering
    \includegraphics[width=0.95\textwidth]{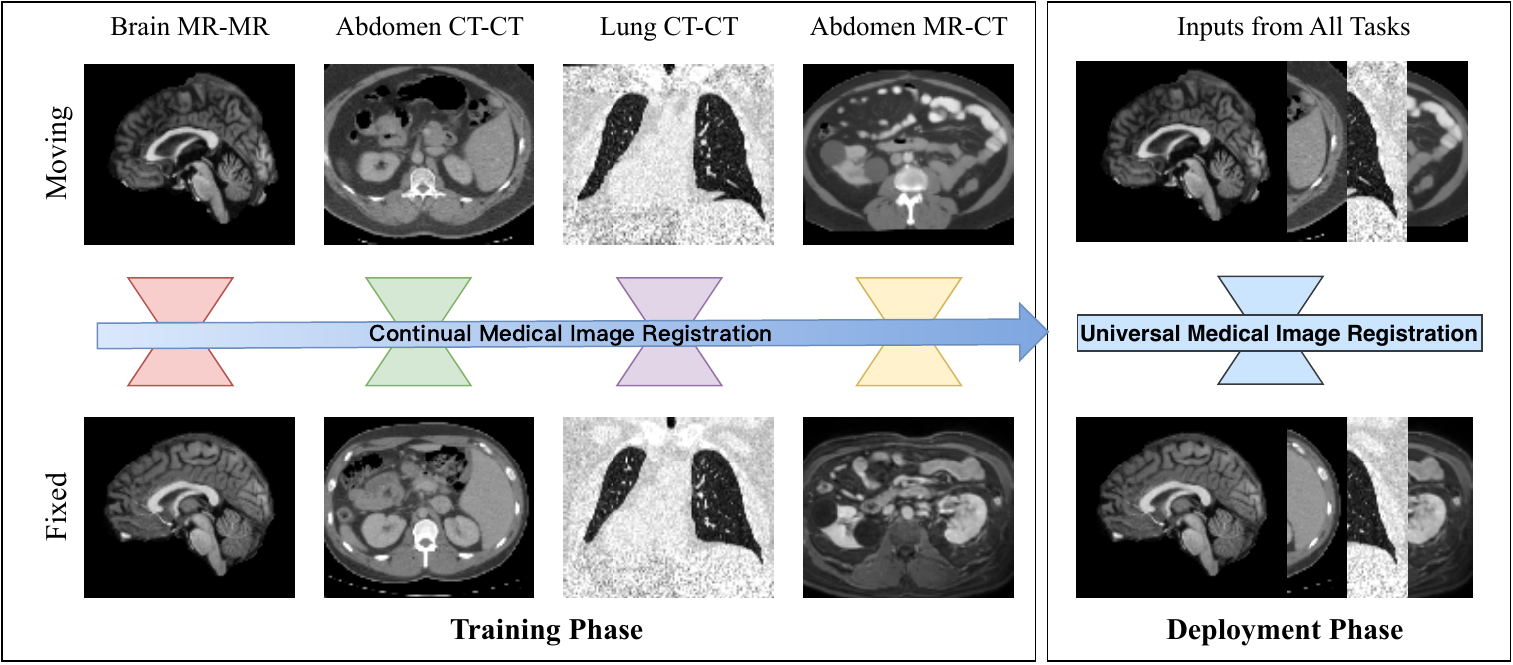}
    \caption{Continual image registration. One can see that the image pairs differ significantly in appearance and anatomy across tasks, \emph{i.e.}, serious distribution shift, posing a major challenge to modern learning-based registration methods.}
    \label{fig:overview}
\end{figure}

\textbf{Toward Universal Medical Image Registration via Continual Learning.}
In stark contrast to current task-specific DL algorithms, \emph{universal medical image registration} is a general-purpose AI for medical imaging, which performs registration on a wide range of clinically relevant tasks simultaneously while maintaining its generalizability to unseen data, and thus has huge potential to be applied to real-world clinics.
To tackle the above challenges, we present the first attempt to achieve the goal of universal 3D medical image registration by proposing a CL method. 
The proposed method continually updates the registration network to adapt to new clinical environments and data and retain the registration performance of old tasks at the same time. 
After sequential training on all tasks encountered, the resulting network can perform well on all tasks. 
Taking four tasks as an example, we illustrate the problem considered in \cref{fig:overview}.  
Specifically, We utilize meta-learning with experience replay to mitigate the problem of catastrophic forgetting. To promote the generalizability of meta-continual learning, we further propose sharpness-aware meta-continual learning (SAMCL) to seek a flat loss landscape. Empirical results on four medical image registration datasets show the effectiveness of our proposed method.

\section{Methodology}
Image registration finds the spatial transformation $\phi:\mathrm{\Omega}_F\rightarrow\mathrm{\Omega}_M$ that registers the moving image $m:\mathrm{\Omega}_M\rightarrow\mathds{R}$ to the fixed image $f:\mathrm{\Omega}_F\rightarrow\mathds{R}$.
Deep learning-based registration methods parameterize the spatial transformation by a neural network taking the image pair as the input with parameters $\bm{\theta}$, \emph{i.e.}, $\phi_{\bm{\theta}}=g(f,m;\bm{\theta})$.
Registration is achieved by optimizing the empirical risk, \emph{i.e.},
\begin{equation}
    \widehat{\bm{\theta}} = \argmin_{\bm{\theta}}
    \mathbb{E}_{(f,m)\sim\mathcal{D}}\left[
    \mathcal{L}(\phi_{\bm{\theta}};f,m)
    \right],
\end{equation}
where $\mathcal{D}=\{(f_i,m_i)\}_{i=1}^{N}$ is a training dataset drawn \emph{i.i.d.} from distribution $\mathcal{P}$, and $\mathcal{L}(\phi_{\bm{\theta}};f,m)=D(f,m\circ\phi_{\bm{\theta}})+ R(\phi_{\bm{\theta}})$ is the loss function composed of a dissimilarity measure $D(\cdot,\cdot)$ and a regularization term $R(\cdot)$.

Collecting a centralized multi-task dataset for training a universal registration network is cumbersome and unrealistic in practice.
In the context of lifelong or \emph{continual} learning (CL) \cite{conference/neurips/thrun1995}, we train the registration network sequentially on non-stationary datasets $\{\mathcal{D}^{(t)}\}_{t=1}^T$, while achieving universal image registration.
This objective can be formulated into the following optimization problem 
\begin{equation}\label{eq:agg_opt}
  \widehat{\bm{\theta}}^{(K)} = \argmin_{\bm\theta}\sum_{t=1}^K \mathbb{E}_{(f,m)\sim\mathcal{D}^{(t)}}\left[\mathcal{L}^{(t)}({\phi}_{\bm{\theta}};f,m)\right], \quad\forall\,K\in\{1,\dots,T\}
\end{equation}
where $\mathcal{L}^{(t)}$ is the loss function for the $t$-th dataset, and $\widehat{\bm{\theta}}^{(K)}$ are the optimal network parameters for the $K$ tasks that have been observed.
When the model learns from one individual dataset to another, it inevitably encounters the issue of generalization toward new datasets and the forgetting of knowledge on previous datasets due to distribution shifts. We propose the method SAMCL to alleviate forgetting of the registration network while promoting its generalizability.

\subsection{Sharpness-Aware Meta-Continual Learning (SAMCL)}
This bi-directional temporal perspective between the past and the future has motivated using meta learning as a means to achieve CL, \emph{a.k.a.} meta-continual learning (Meta-CL).
Meta-CL relies on rehearsal that replays previous task samples using a small memory buffer and encourages transfer between past and current samples.
For instance, Meta Experience Replay (MER) \cite{conference/iclr/riemer2019} exploits the Reptile algorithm \cite{journal/arxiv/nichol2018}, which essentially optimizes the following objective:
\begin{equation*}
  \min_{\bm\theta}\mathbb{E}_{(f_1,m_1),\dots,(f_b,m_b)\sim\mathcal{M}}
  \left[
    \sum_{i=1}^b\left(
    \mathcal{L}^{(t_i)}(\phi_{\bm{\theta}}) - \frac{\alpha}{2} 
    \sum_{j=1}^{i-1}
    \nabla_{\bm{\theta}}\mathcal{L}^{(t_i)}(\phi_{\bm{\theta}})\cdot
    \nabla_{\bm{\theta}}\mathcal{L}^{(t_j)}(\phi_{\bm{\theta}})
    \right)
  \right]
\end{equation*}
where $\cdot$ is the inner product, $\mathcal{M}$ is a memory buffer of size $b$ acquired by Experience Replay (ER) \cite{conference/neurips/rolnick2019}, and $\mathcal{L}^{(t_i)}$ is the registration loss for the image pair $(f,m)\sim\mathcal{D}^{(t_i)}$.
Therefore, MER tries to find a gradient direction that can simultaneously minimize losses on all datasets using samples from the memory buffer.

Nevertheless, since the objective function for deformable image registration is highly \emph{non-convex and there may exist numerous local minima} \cite{journal/pmb/hill2001}, leading to poor generalizability.
Besides, its gradient \emph{w.r.t.} the network parameters can be very noisy and may not represent a direction minimizing the expected loss $\mathbb{E}_{(f,m)\sim\mathcal{P}}\left[\mathcal{L}(\phi_{\bm{\theta}};f,m)\right]$.
Inspired by the recent progress in sharpness-aware minimization (SAM) which improves the generalizability of neural networks \cite{conference/iclr/foret2021}, we hypothesize that integrating SAM into Meta-CL could enhance the performance on continual image registration.
This new approach optimizes the objective
\begin{equation}  
\min_{\bm\theta}\mathbb{E}_{(f_1,m_1),\dots,(f_b,m_b)\sim\mathcal{M}}
  \left[
    \sum\limits_{i=1}^b\left(
    \mathcal{L}_{\text{sam}}^{(t_i)}(\phi_{\bm{\theta}}) - \frac{\alpha}{2} 
    \sum\limits_{j=1}^{i-1}
    \nabla_{\bm{\theta}}\mathcal{L}_{\text{sam}}^{(t_i)}(\phi_{\bm{\theta}})\cdot \nabla_{\bm{\theta}}\mathcal{L}_{\text{sam}}^{(t_j)}(\phi_{\bm{\theta}})
    \right)
  \right]
\end{equation}
where $\mathcal{L}_{\text{sam}}^{(t_i)}(\phi_{\bm{\theta}})\triangleq \max_{\bm{\epsilon}:\norm{\bm{\epsilon}}_2\leq\rho} \mathcal{L}^{(t_i)}(\phi_{\bm{\theta}+\bm{\epsilon}})$ and $\rho \geq 0$ is a hyperparameter. The motivation of SAM is to seek parameters that lie in neighborhoods having uniformly low loss value
(rather than parameters that only themselves have low loss value), thus simultaneously minimizing loss value and loss sharpness. To solve the min-max optimization problem, the inner maximization problem is first approximated via a first-order Taylor expansion of $\mathcal{L}^{(t_i)}(\phi_{\bm{\theta}+\bm{\epsilon}})$ around $\bm{\epsilon}=\bm{0}$. By solving a classical dual norm problem\cite{conference/iclr/foret2021}, we can obtain the solution

\begin{equation}
\begin{aligned}
  \widehat{\bm{\epsilon}}(\bm{\theta})\triangleq
  \argmax_{\bm{\epsilon}:\norm{\bm{\epsilon}}_2\leq\rho}\mathcal{L}^{(t_i)}(\phi_{\bm{\theta}+\bm{\epsilon}})
  &\approx
  \argmax_{\bm{\epsilon}:\norm{\bm{\epsilon}}_2\leq\rho}\left\{\mathcal{L}^{(t_i)}(\phi_{\bm{\theta}}) + \bm{\epsilon}^{\intercal}\nabla_{\bm{\theta}}\mathcal{L}^{(t_i)}(\phi_{\bm{\theta}})\right\} \\
  &=\argmax_{\bm{\epsilon}:\norm{\bm{\epsilon}}_2\leq\rho}\bm{\epsilon}^{\intercal}\nabla_{\bm{\theta}}\mathcal{L}^{(t_i)}(\phi_{\bm{\theta}})
  = \rho \frac{\nabla_{\bm{\theta}}\mathcal{L}^{(t_i)}(\phi_{\bm{\theta}})}{\norm{\nabla_{\bm{\theta}}\mathcal{L}^{(t_i)}(\phi_{\bm{\theta}})}_2}.
\end{aligned}
\end{equation}

As SAM seeks for flat local minima, there would be increasing possibilities that the gradient $\nabla_{\bm{\theta}}\mathcal{L}_{\text{sam}}^{(t_i)}(\phi_{\bm{\theta}})\approx\nabla_{\bm{\theta}}\mathcal{L}^{(t_i)}(\phi_{\bm{\theta}})\big|_{\bm{\theta}+\widehat{\bm{\epsilon}}(\bm{\theta})}$ corresponds to the direction that minimizes the expected loss with $\widehat{\bm{\epsilon}}(\bm{\theta})$.
\cref{alg:samcl} summarizes the proposed algorithm for sharpness-aware meta-continual learning.

\begin{algorithm}[t]
  \caption{Sharpness-Aware Meta-Continual Learning (SAMCL)}
  \KwData{Training image pairs $\mathcal{D}^{(t)}=\{(f_i,m_i)\}_{i=1}^{N_t}$, $t=1,\dots,T$\;}
  \KwIn{Learning rate $\alpha$, meta-learning rate $\beta$, neighbourhood size $\rho$, batch size $s$\;}
  \KwOut{Trained network parameters $\widehat{\bm\theta}^{(T)}$\;}
  \textbf{Initialization}: randomly sample $\widehat{\bm{\theta}}^{(0)}$; set memory buffer $\mathcal{M}=\emptyset$\;
  \For{$t=1,\dots,T$}{
    Initialize batches $\mathcal{B}_t=\{(f_i,m_i)\}_{i=1}^{b_t}$; initialize  $\bm{\theta}\leftarrow\widehat{\bm{\theta}}^{(t-1)}$\;
    \For{$i=1,\dots,b_t$}{
      $\bm{\theta}_0\leftarrow\bm{\theta}$; $B_i\leftarrow(f_i,m_i)$\;
      \For{$j=1,\dots,s$}{
        Compute $\widehat{\bm{\epsilon}}(\bm{\theta})= \nicefrac{\rho\nabla_{\bm{\theta}}\mathcal{L}^{(t)}(\phi_{\bm{\theta}};B_i[j])}{\norm{\nabla_{\bm{\theta}}\mathcal{L}^{(t)}(\phi_{\bm{\theta}};B_i[j])}_2}$\;
        Update parameters $\bm{\theta}\leftarrow\bm{\theta}-\alpha\nabla_{\bm{\theta}}\mathcal{L}^{(t)}(\phi_{\bm{\theta}};B_i[j])\big|_{\bm{\theta}+\widehat{\bm{\epsilon}}(\bm{\theta})}$\;
      }
      $\mathcal{M}\leftarrow\mathcal{M}\cup\{(f_i,m_i)\}$\tcp*{Reservoir sampling memory update}
      $B_i\leftarrow\operatorname{sample}(\mathcal{M})$\tcp*{sample from the memory buffer}
      Compute $\widehat{\bm{\epsilon}}(\bm{\theta})= \nicefrac{\rho\nabla_{\bm{\theta}}\mathcal{L}^{(t_i)}(\phi_{\bm{\theta}};B_i)}{\norm{\nabla_{\bm{\theta}}\mathcal{L}^{(t_i)}(\phi_{\bm{\theta}};B_i)}_2}$\;
      Update parameters $\bm{\theta}\leftarrow\bm{\theta}-\alpha\nabla_{\bm{\theta}}\mathcal{L}^{(t_i)}(\phi_{\bm{\theta}};B_i)\big|_{\bm{\theta}+\widehat{\bm{\epsilon}}(\bm{\theta})}$\;$\bm{\theta}\leftarrow\bm{\theta}+\beta(\bm{\theta}-\bm{\theta}_0)$\tcp*{meta-learning update}
      
    }
    $\widehat{\bm{\theta}}^{(t)}\leftarrow\bm{\theta}$\;
  }
  \Return{$\widehat{\bm{\theta}}^{(T)}$.}
  \label{alg:samcl}
\end{algorithm}

\subsubsection{Toward Universal Medical Image Registration.}
We achieve universal 3D medical image registration via continual learning on sequentially observed datasets.
We use the Large Kernel U-Net (LK-UNet) as the backbone of the registration network with 16 kernels in its first layer, which performed well on several registration tasks \cite{conference/mlmi/jia2022}.
The objective of proposed SAMCL is to retain the registration accuracy on observed datasets after the network is sequentially trained by all the tasks, as well as improve its generalizability on unseen domains.

\section{Experiments and Results}

\begin{table}[t]
  \centering
  \caption{Evaluation metrics for universal medical image registration on the OASIS/Abdomen CT-CT/NLST/Abdomen MR-CT datasets, respectively.
  The results for NLST in terms of TRE are marked in italics.
  Better performance of SAMCL than other methods is indicated by underlining.
  Note that the NLST data were evaluated by TREs in millimeters.
  Statistical significant difference indicated by a paired $t$-test for $p<0.05$ was marked by asterisks.}
  {% This command increases the font size of the table
  \begin{tabular}{L{2cm}@{\hspace{-0.5ex}}C{5cm}@{\hspace{-0.5ex}}C{5cm}}
    \toprule
    \textbf{Method} & \textbf{AVG Dice$\uparrow$/\textit{TRE} (mm)$\downarrow$} & \textbf{BWT Dice$\uparrow$/\textit{TRE} (mm)$\downarrow$} \\
    \midrule
    None & \underline{0.598}*/\underline{0.311}*/\underline{\textit{7.932}}*/\underline{0.353}* & --- \\
    Independent & 0.738*/\underline{0.335}*/\textit{3.259}*/\underline{0.550} & --- \\
    Multi-task & 0.743*/0.358/\textit{2.662}*/\underline{0.506}  & --- \\
    \midrule
    Sequential & \underline{0.348}*/\underline{0.270}/\underline{\textit{6.922}}*/0.568 & \underline{-0.395}*/\underline{-0.129}*/\underline{\textit{3.508}}*/0 \\
    EWC \cite{journal/pnas/kirkpatrick2016} & \underline{0.659}*/0.364*/\textit{3.836}/\underline{0.506}  & \underline{-0.083}*/-0.011*/\underline{\textit{0.386}}/0 \\
        SI \cite{si_zenke2017continual} & \underline{0.461}*/\underline{0.303}*/\underline{\textit{8.186}}*/\underline{0.555}  & \underline{-0.281}*/\underline{-0.084}*/\underline{\textit{4.685}}*/0 \\ 
    GPM \cite{conference/iclr/saha2021} & \underline{0.458}*/\underline{0.290}*/\underline{\textit{7.095}}*/\underline{0.556}  & \underline{-0.281}*/\underline{-0.085}*/\underline{\textit{3.574}}*/0 \\ 
    MER \cite{conference/iclr/riemer2019} & 0.698/\underline{0.331}*/\textit{3.611}*/\underline{0.532}  & -0.036/\underline{-0.040}/\textit{0.077}*/0 \\ 
    \hdashline\noalign{\vskip 0.5ex}
    SAMCL & 0.697/0.352/\textit{3.855}/0.563  & -0.037/-0.039/\textit{0.306}/0 \\ 
    \bottomrule
  \end{tabular}}
  \label{tab:result}
\end{table}

\subsubsection{Materials and Compared Methods.}
The experiments were performed on four sequentially arrived datasets from the Learn2Reg challenges \cite{journal/tmi/hering2021}, in order of OASIS (brain MRI, inter-subject) \cite{journal/jcn/marcus2007}, Abdomen CT-CT (inter-subject) \cite{journal/tbe/xu2016}, NLST (lung CT, intra-subject) \cite{journal/jdi/clark2013}, and Abdomen MR-CT (intra-subject) \cite{journal/jdi/clark2013,journal/tbe/xu2016}.
The tasks were assigned as in \cref{fig:overview}, varying in image modalities and ROIs.
All image volumes were resampled to spacing of $2\times 2\times 2$ mm for the OASIS and $3\times 3\times 3$ mm for the other datasets, and then cropped/padded to size of $112\times 96\times 112$.
The number of training/validation/test images for each task is 290/40/84 (OASIS), 21/3/6 (Abdomen CT-CT), 294/42/84 (NLST), and 80/9/16 (Abdomen MR-CT).
The intensity range of CT images was clipped to $[-200,300]$ to cover abdominal organs.
All images were linearly normalized to [0,1] as network input (for CT images this was after intensity clipping).
To avoid different memory buffers and promote fair comparison across compared methods, no data augmentation was performed on the training set.
We compared the proposed SAMCL with different baselines, including
(i) independent learning on each task using different models,
(ii) multi-task learning on all tasks using a single model, 
(iii) vanilla sequential learning over all tasks using a single model, and
(iv) previous CL methods EWC
\cite{journal/pnas/kirkpatrick2016}, SI
\cite{si_zenke2017continual},
GPM \cite{conference/iclr/saha2021}, and MER \cite{conference/iclr/riemer2019}. Note that independent and multi-task learning can be the upper bounds for task-specific performance and average performance of all tasks, respectively.

\subsubsection{Evaluation Metrics.}
The quality of registration was evaluated by anatomical labels or landmarks.
For the OASIS, Abdomen CT-CT and MR-CT datasets, the average Dice similarity coefficient (DSC) was computed on 35, 13, and 9 corresponding labels, respectively.
For the NLST dataset, we use the target registration error (TRE) in millimeters on corresponding landmarks as the evaluation metric. 
Denoting $R_{i,j}$ as the average metric evaluated on task $j$ by the model trained right after task $i$, we report the average performance (AVG) and backward transfer (BWT) defined as 
\begin{equation}
  \operatorname{AVG}_i\triangleq R_{T,i},\qquad 
  \operatorname{BWT}_i\triangleq R_{T,i}-R_{i,i}.
\end{equation}
$\operatorname{AVG}_i$ refers to the performance of the registration network on test images of task $i$ after learning all tasks, while $\operatorname{BWT}_i$ signifies the forgetting of task $i$, i.e., the performance decay caused by subsequent tasks.

\subsubsection{Implementation Details.}
Experiments were conducted using PyTorch on an NVIDIA RTX$^{\text{TM}}$ 3090 GPU. 
For each task, training was performed for 10000 iterations by the Adam optimizer \cite{conference/iclr/kingma2014} with a learning rate of $1\times 10^{-4}$. 
The batch size was set to 4. The memory buffer size of SAMCL was set to 200 while the meta-learning rate was chosen as 0.25. The $\rho$ parameter of SAM was set to 0.05 as it produced the best performance. 
Besides, since the images from different tasks are annotated by different labels, the loss function for each task also varies.
Specifically, for the OASIS, Abdomen CT-CT and Abdomen MR-CT tasks the loss function consists of a negative local normalized cross-correlation (LNCC) with window size of 3 \cite{journal/media/avants2008} and a soft dice loss \cite{journal/media/hu2018}, while for NLST images the loss function comprises a negative LNCC and a landmark distance loss based on TRE.
The registration is performed symmetrically between the fixed and moving images using a diffeomorphic transformation parameterized by stationary velocity fields.
A regularization term composed of membrane and bending energy is also included to enforce deformation smoothness \cite{journal/ni/ashburner2007}, whose weights were set to $1$ during training.

\subsubsection{Comparison Study.}

\cref{tab:result} presents the results of the comparison study.
In terms of average registration accuracy (AVG Dice/TRE), SAMCL outperformed previous CL methods like EWC, SI and GPM on at least one of the four tasks.
Note that for the Abdomen CT-CT task, both independent and multi-task learning can only achieve minor improvement due to the scarcity of training data and large inter-subject deformations.
In terms of backward transfer (BWT Dice/TRE) which indicates the stability and robustness to knowledge forgetting of a model, SAMCL achieved significant improvement over sequential training and performed better than or on par with previous CL methods, which demonstrates its advantage in a continual learning setup. 

\begin{figure}[t]
    \centering
    \includegraphics[width=1\textwidth]{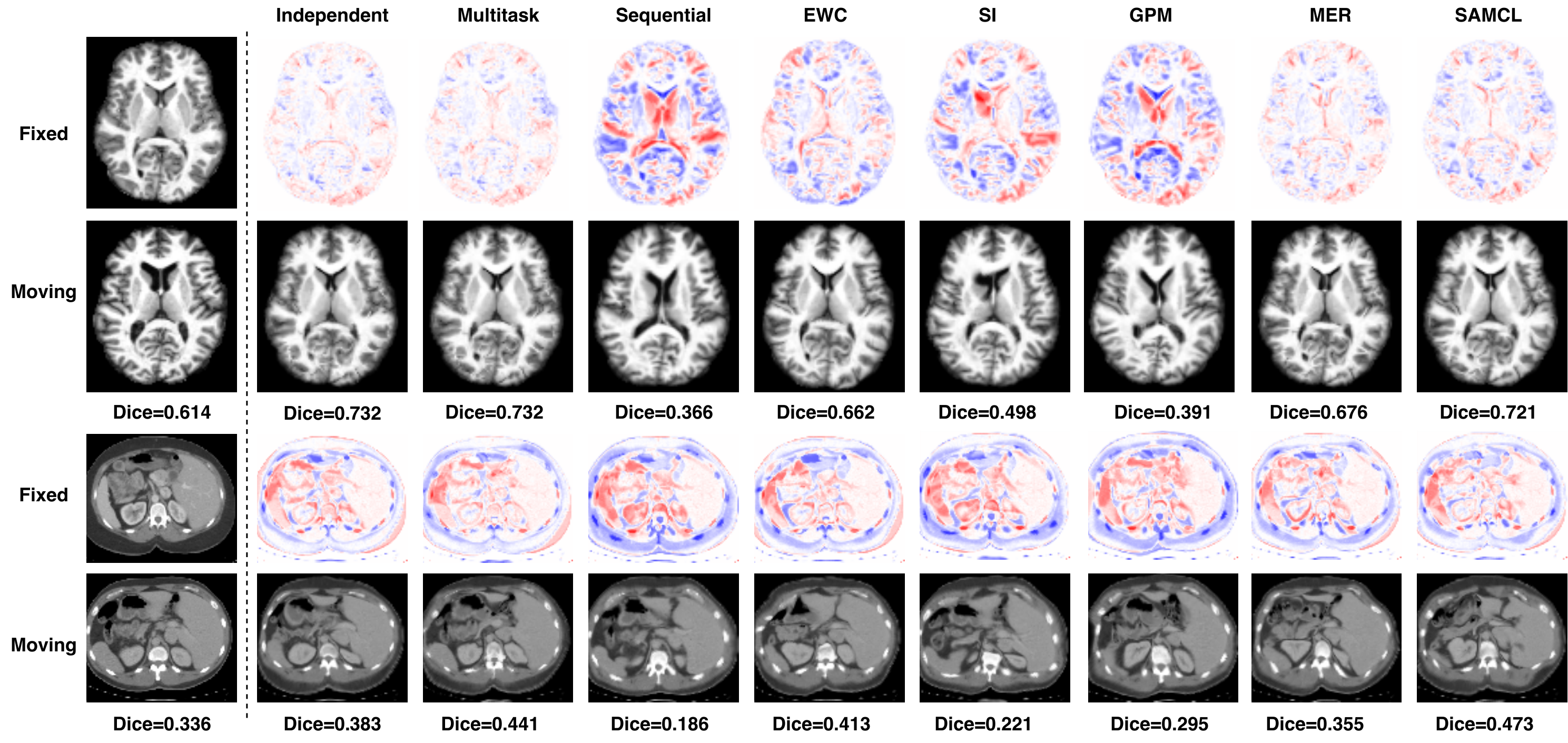}
    \caption{Registration results on an exemplar OASIS and Abdomen CT image pair using different training strategies. 
    The top row visualizes the difference map between the fixed image and the registered moving image.
    Note that CT intensities were clipped to the range $[-200, 300]$.}
    \label{fig:vis}
\end{figure}

Furthermore, SAMCL performed comparable to independent training and centralized multi-task learning, indicating its potential for universal image registration in a real-world clinical environments.
\cref{fig:vis} visualizes an exemplar registered image pair from the OASIS and Abdomen CT data using different training strategies, where SAMCL outperformed previous CL methods and achieved comparable performance to centralized multi-task training. Please refer to the supplementary material for more visualization of the results.

\subsubsection{Ablation and Generalization Study.}

\begin{figure}[t]
    \centering
    \begin{subfigure}{0.45\textwidth}
        \includegraphics[width=\textwidth]{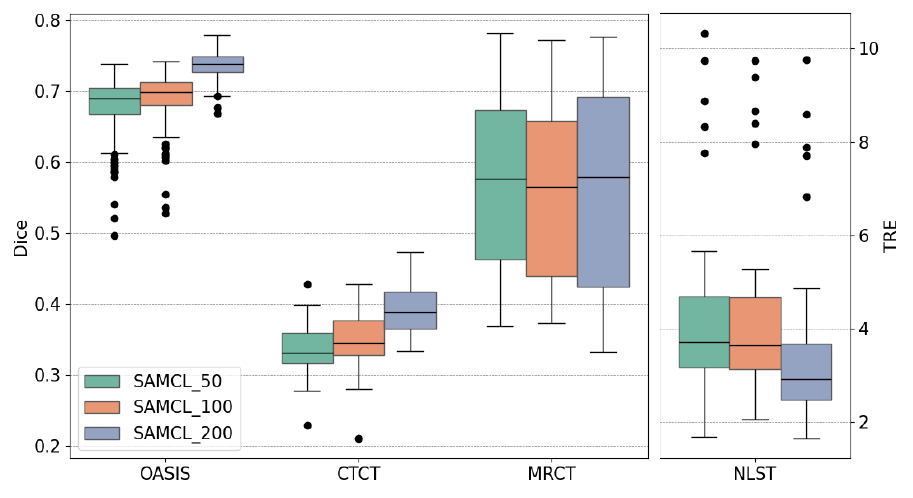}
        \caption{Ablation study.}
        \label{fig:ablation}
    \end{subfigure}
    \begin{subfigure}{0.52\textwidth}
        \includegraphics[width=\textwidth]{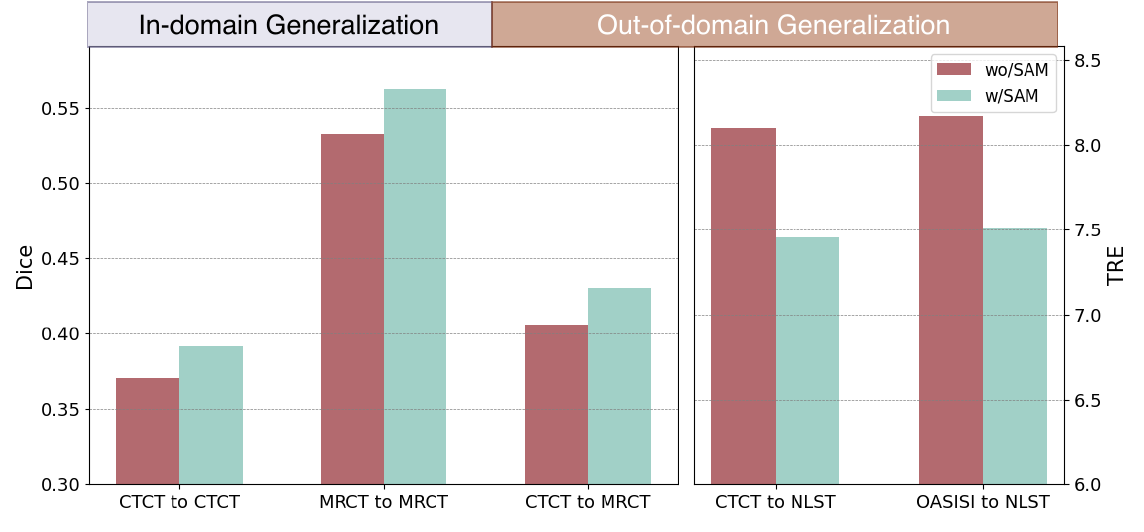}
        \caption{Generalization study.}
        \label{fig:generalization}
    \end{subfigure}
    \caption{Ablation and generalization study of the proposed SAMCL on (a) using various memory buffer sizes for experience replay, and (b) using sharpness-aware minimization (SAM) for in- and out-of-domain generalization tasks.}
\end{figure}

\cref{fig:ablation} performs the ablation study on using different memory buffer sizes for our proposed SAMCL.
One can observe that for the first three tasks (OASIS, Abdomen CT-CT and MR-CT), increasing buffer size has a positive effect on the registration accuracy.
This is because using a larger memory size improves the preservation of knowledge on previous tasks.
We also investigate the effectiveness of the sharpness-aware minimization (SAM) in terms of its performance gains for in- and out-of-domain generalization, where a model trained by early tasks is tested on unseen data from later tasks.
\cref{fig:generalization} presents the bar plots for different generalization tasks with or without SAM.
The result shows that for both in-domain (Abdomen CT-CT/MR-CT) and out-of-domain (Abdomen CT-CT to MR-CT, Abdomen CT-CT to NLST, OASIS to NLST) generalization tasks, our model equipped with SAM outperforms the vanilla Meta-CL without SAM, highlighting the advantage of SAMCL as a training strategy for simultaneously improving model generalizability.

\section{Conclusion and Discussion}
In this paper, we introduce the first effort toward accomplishing the goal of universal 3D medical image registration through a continual learning approach. We utilize meta-learning with experience replay to tackle the catastrophic forgetting problem. We propose to use sharpness-aware training to enhance the generalization performance. 
Empirical results show that the proposed method can achieve promising registration performance, comparable to centralized multitask and independent training strategies. 
The sequence in which tasks are presented to the model may affect the universal registration performance. In future work, we will evaluate the task order robustness of SAMCL with more datasets from different modalities and anatomies. 

\begin{figure}
  \centering
  \includegraphics[width=\textwidth]{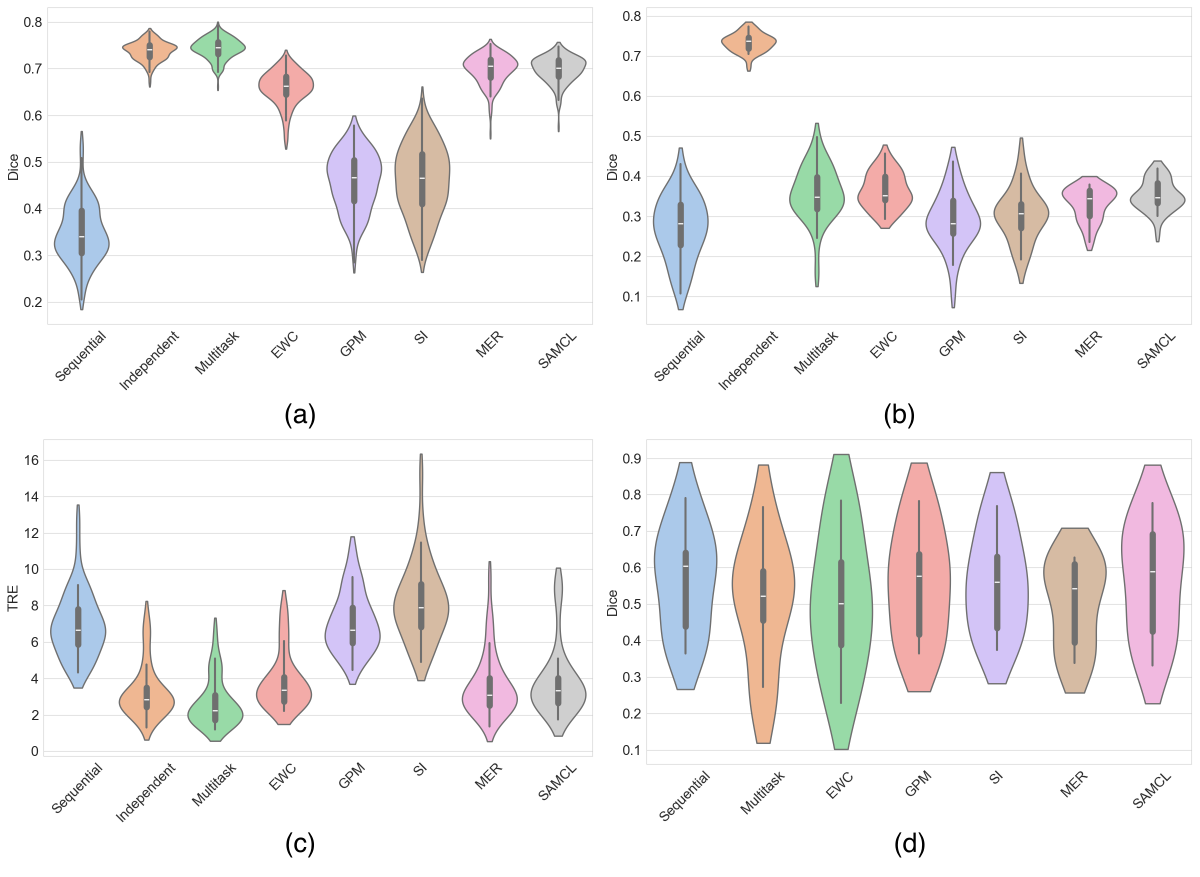}
  \caption{Violin plot to show the registration performance of different methods on (a)OASIS, (b)Abdomen CT-CT, (c)NLST and (d) Abdomen MR-CT.}
\end{figure}

\begin{figure}
  \centering
  \includegraphics[width=\textwidth]{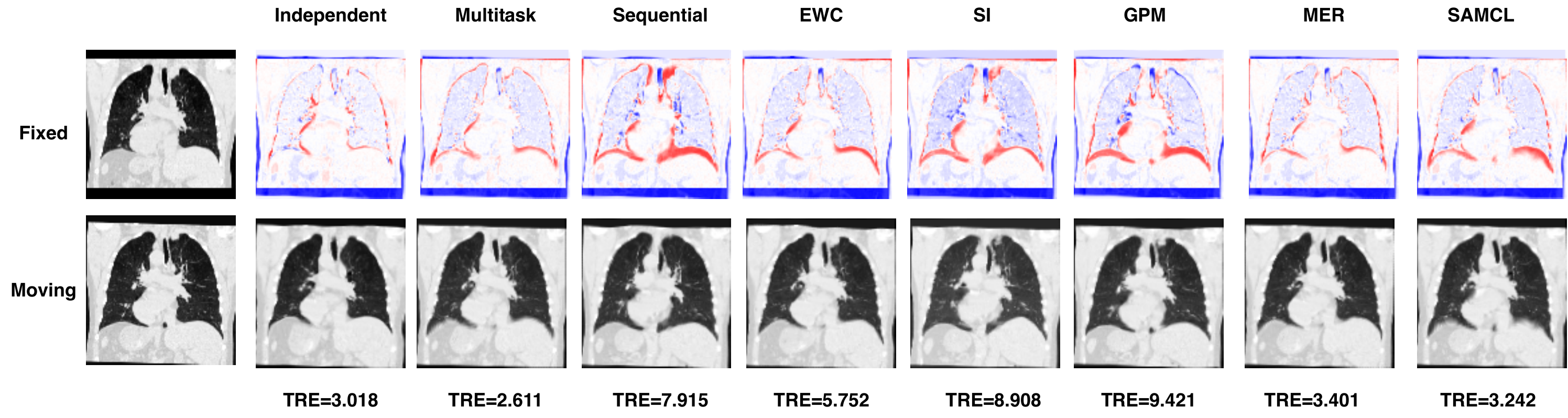}
  \caption{Registration results on an exemplar NLST image pair using different training strategies. 
  The top row visualizes the difference map between the fixed image and the registered moving image.}
\end{figure}

\begin{credits}
\subsubsection{\ackname}
This work was funded by the National Natural Science Foundation of China (grant No. 62372115, 61971142 and 62111530195).
    
\subsubsection{\discintname}
The authors have no competing interests to declare that are relevant to the content of this article.
\end{credits}

\bibliographystyle{splncs04}
\bibliography{main}

\end{document}